\documentclass[runningheads]{llncs}
\usepackage[T1]{fontenc}
\usepackage{graphicx}
\usepackage{booktabs}
\usepackage{multirow}
\usepackage[normalem]{ulem}
\usepackage[table]{xcolor}
\renewcommand{\arraystretch}{1.2}
\useunder{\uline}{\ul}{}
\usepackage{hyperref}
\usepackage{orcidlink}

\begin{document}
\title{Leveraging Multi-View Weak Supervision for Occlusion-Aware Multi-Human Parsing}
\titlerunning{Multi-View Weak Supervision for Occlusion-Aware Multi-Human Parsing}
%
\author{Laura Bragagnolo\inst{1}\orcidlink{0009-0007-8096-4588} \and
Matteo Terreran\inst{1}\orcidlink{0000-0001-9862-8469} \and
Leonardo Barcellona \inst{2}\orcidlink{0000-0003-4281-0610} \and
Stefano Ghidoni \inst{1}\orcidlink{0000-0003-3406-8719}}
\authorrunning{L. Bragagnolo et al.}
%
\institute{Department of Information Engineering, University of Padova, Italy \and
University of Amsterdam, The Netherlands \\
\email{laura.bragagnolo.1@phd.unipd.it}}
\maketitle              
\begin{abstract}
    Multi-human parsing is the task of segmenting human body parts while associating each part to the person it belongs to, 
    combining instance-level and part-level information for fine-grained human understanding.
    In this work, we demonstrate that, while state-of-the-art approaches achieved notable results on public datasets, they struggle considerably in segmenting people with overlapping bodies.
    From the intuition that overlapping people may appear separated from a different point of view, we propose a novel training framework exploiting multi-view information to improve multi-human parsing models under occlusions.
    Our method integrates such knowledge during the training process, introducing a novel approach based on weak supervision on human instances and a multi-view consistency loss.
    Given the lack of suitable datasets in the literature, we propose a semi-automatic annotation strategy to generate human instance segmentation masks from multi-view RGB+D data and 3D human skeletons.
    The experiments demonstrate that the approach can achieve up to a 4.20\% relative improvement on human parsing over the baseline model in occlusion scenarios.
\end{abstract}
\section{Introduction}
Multi-human parsing (MHP), also known as instance-level human parsing, extends human body parts segmentation by integrating information about the specific person the part belongs to.
It finds many applications in the fields of multimedia and computer vision, such as human pose estimation~\cite{guler2018densepose} and fashion image manipulation~\cite{dong2020fashion}.

While current multi-human parsing methods achieve impressive results~\cite{zhang2022aiparsing,cheng2022masked}, the task remains highly challenging. One difficulty is the lack of a standardized set of semantic labels across datasets. For example, the PASCAL-Person-Part dataset~\cite{chen2014detect} considers only 6 body limb classes, while the widely used Crowd Instance-level Human Parsing (CIHP) dataset~\cite{gong2018instance} takes into account 19 labels, including clothing and accessories. Such  granularity has led to the development of domain-specific approaches. 
But the most critical challenge arises from occlusions, particularly those caused by overlapping bodies~\cite{cheng2022masked}. To quantify the effect of these, in this work, we analyze the impact of varying levels of occlusion on the accuracy of existing multi-human parsing models. 
We find that current methods struggle to accurately detect body parts when people are occluding each other, an event that scales with the number of people in the scene. Large portions of the body may be entirely missed, and body parts are frequently mislabeled or assigned to the wrong person.

In this work, we show that multiple viewpoints offer a chance of effectively recovering information about human shapes: overlapped bodies in one image may be  separated in a different viewpoint.
Building on this idea, we propose a novel training framework to leverage multi-view information to fine-tune MHP models.
Our approach is based on two main elements:
(i) single-view loss on shape information for weak supervision and
(ii) a multi-view instance and body part consistency loss. The former provides coarse-grained information about human shapes, while the latter ensures coherent identity and part predictions across multiple views. The combination of the two is referred to as Multi-View Instance-Guided Multi-Human Parsing (MVIG-MHP) framework, while Instance-Guided Multi-Human Parsing (IG-MHP) represents the framework comprising of the single-view contributions only.
Our frameworks can be applied to any MHP model, trained on a standard MHP dataset, to improve its performance on occlusions.

As the literature lacks a multi-view, multi-human instance segmentation dataset that captures occlusions between people, we propose a new semi-automatic annotation pipeline that generates human instance segmentation masks from multi-view RGB+D data and 3D human skeletons.
The pipeline is applied to video sequences displaying multiple people from the CMU Panoptic Studio dataset~\cite{Joo_2017_TPAMI}, resulting in a new dataset named Panoptic-HuIS. Unlike annotations from multi-human parsing ground truth, the dataset only considers human shape, providing coarser labels, making our framework applicable to models pre-trained on different MHP datasets, independent from their body-part label convention and granularity.
 
\begin{figure}[tb]
    \centering
    \includegraphics[width=0.8\textwidth]{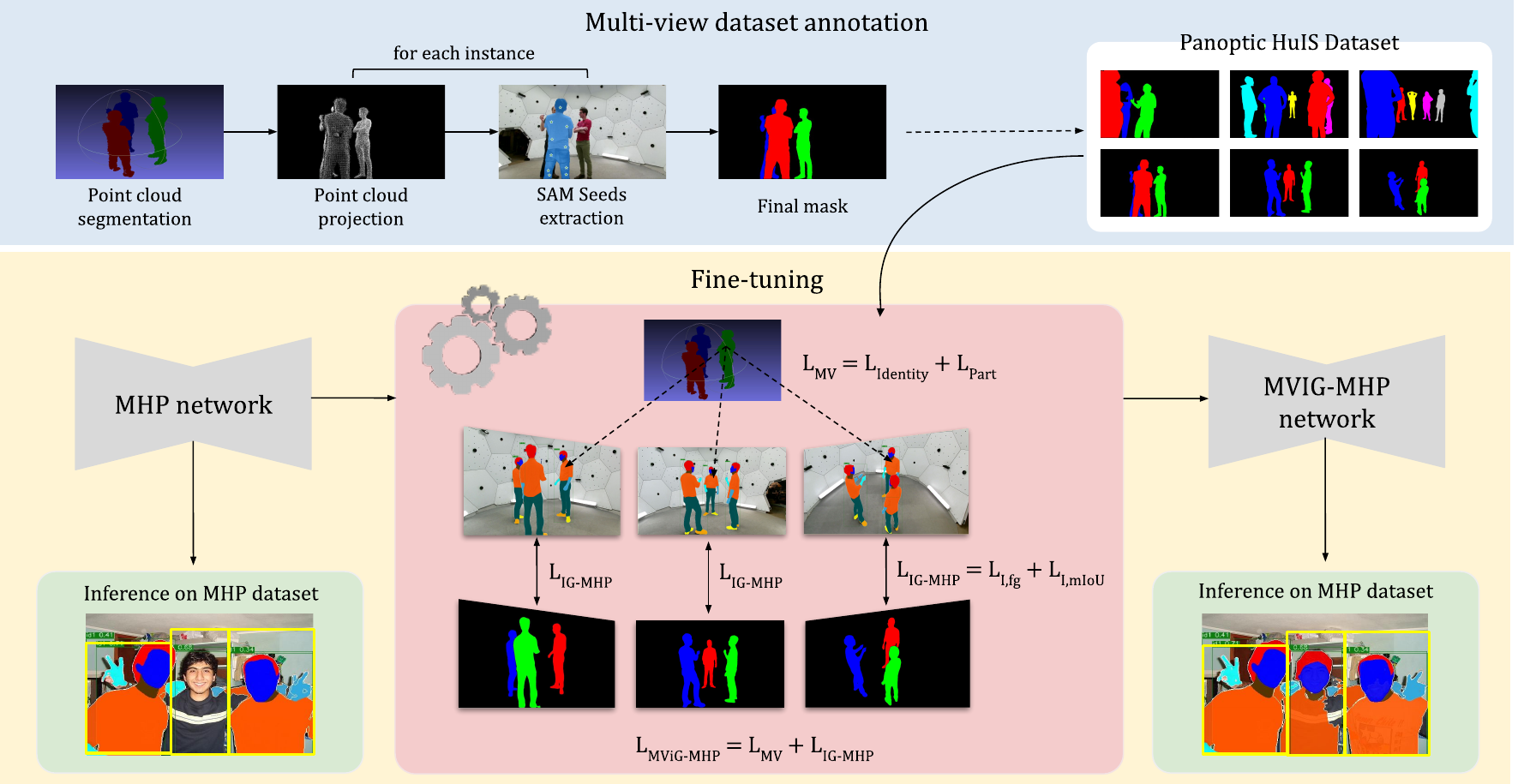}
    \caption{Our approach for occlusion-robust multi-human parsing includes a multi-view dataset annotation pipeline (top), generating human segmentation masks for the fine-tuning process (middle). The inference phase (bottom, in green) shows results before and after applying MVIG-MHP.}
    \label{fig:framework}
\end{figure}

Figure~\ref{fig:framework} provides an overview of the proposed approach. The upper section illustrates the multi-view dataset annotation process, generating human instance segmentation masks used in the fine-tuning process of a multi-human parsing model through two proposed stages, combined into the MVIG-MHP framework.

To summarize, the main contributions of this paper are:
(i) a comprehensive analysis of the impact of occlusions on state-of-the-art MHP models; (ii) Panoptic-HuIS, a new dataset including scenes with many people overlapping, with human instance segmentation annotations; (iii) a training framework exploiting Panoptic-HuIS to enhance the performance of pretrained MHP models in cases of significant overlaps, thanks to a novel multi-view consistency loss; (iv) evaluation demonstrating the effectiveness in improving on the CHIP dataset when people are overlapping.

\section{Related Works}

\subsubsection{Multi-human parsing.}
Multi-human parsing methods can be categorized into three main classes: bottom-up, top-down, and single-stage approaches.
Bottom-up methods first predict pixel-level body-part labels, then group them into human instances. PGN~\cite{gong2018instance} segments body parts and then uses edge detection for grouping, while M2FP~\cite{yang2023deep}, building on the Mask2Former architecture~\cite{cheng2022masked}, exploits the hierarchical structure of the human body and self-attention to relate body parts and instances.

Top-down methods follow a different approach by first detecting individuals and then segmenting body parts for each human instance. These techniques are further divided into two-stage and one-stage methods.
The former relies on a robust pre-trained object detector, such as Mask R-CNN~\cite{he2017mask}, to detect human bounding boxes, followed by single-human parsing~\cite{ruan2019devil}.
The latter integrates detection and parsing into a single unified framework.
One example is Parsing R-CNN~\cite{yang2019parsing}, which improves segmentation quality via a geometric and context encoding module that captures relationships between body parts. RP R-CNN~\cite{yang2020renovating} further improves semantics by integrating a semantically enhanced Feature Pyramid Network (FPN)~\cite{lin2017feature}.
Differently from most top-down approaches, AIParsing~\cite{zhang2022aiparsing} utilizes the anchor-free detection head FCOS~\cite{tian2019fcos} to localize human instances and combines a CNN-based backbone, such as~\cite{he2016deep}, with a FPN to extract multi-scale features for parsing. 

Single-stage models directly predict part labels and instances in one pass, avoiding detection or grouping. RepParser~\cite{dai2023resparser} injects instance-aware keypoints in the human parsing pipeline, to dynamically parse each person. SMP~\cite{chu2023single} uses body center features to obtain instance segmentation and generates offsets from the center of the body to the parts to match without any grouping process. Similarly, Uniparser~\cite{chu2024uniparser} proposes a unified correlation representation to learn instance and category features within the cosine space.

\subsubsection{Multi-view human parsing.}
Multi-view data remains largely unexplored in instance-level human parsing, despite its established use in 3D segmentation tasks~\cite{robert2022learning}, including 3D body part segmentation~\cite{terreran2021multi,humancarving2021}.
In~\cite{terreran2021multi} authors propose a multi-view system to create a semantic 3D representation of the human body robust to occlusions, aggregating segmentation from each camera into a point cloud.
Similarly, Li et al.~\cite{humancarving2021} integrate human parsing into multi-view 3D human reconstruction, producing a semantic reconstruction used to estimate human shape parameters.
However, their representations do not distinguish between different human instances. 
In contrast, our method provides fine-grained instance-level information, while handling occlusions and improving segmentation accuracy.
Antonello et al.~\cite{antonello2018multi} present a frame fusion technique to enhance single-view semantic segmentation. Inspired by their approach, we introduce a multi-view loss to enforce instance identity and body part prediction consistency across views.

\section{Method}
\label{sect:Method}
Our novel training framework for MHP consists of a fine-tuning procedure of pre-trained MHP models using two main contributions: (i) single-view loss functions to guide the segmentation of individual instances promoting accurate body part segmentation, 
denoted as Instance-Guided Multi-Human Parsing (IG-MHP); (ii) a multi-view consistency loss, leveraging 3D information to resolve ambiguities caused by overlapping instances and enforcing coherent predictions across views. The combination of both loss contributions, denoted as Multi-View Instance-Guided Multi-Human Parsing (MVIG-MHP), is illustrated in Figure~\ref{fig:framework}.

\subsection{Instance-Guided Multi-Human Parsing}
\label{sect:IG-MHP}
Instance-guided multi-human parsing focuses on guiding the network to accurately segment individual human bodies in a single image. Human instance segmentation annotations are used to train the model using a loss function consisting of two terms: a foreground instance loss ($L_{I,fg}$) and a mean Intersection-over-Union loss ($L_{I,mIoU}$).
The foreground instance loss $L_{I,fg}$ encourages accurate segmentation of the target human instance while treating as background other bodies that are overlapping with it. As a different part-level map is obtained for each different human body, such maps are combined to derive full-body segmentation predictions. Specifically, the union of body part maps is computed by taking the maximum output value for each pixel:
\begin{equation}
        p_{h}(x) = \max \{p_i(x):i \in C\}\,,
\end{equation}
where $C$ represents the set of body part categories.
This loss is implemented as the cross-entropy loss between the segmentation predicted by the network and the human ground truth mask, as defined in Equation~\ref{eq:ce}, where $\mathcal{X}$ represents the set of all pixel coordinates $x$.
\begin{equation}
\label{eq:ce}
        L_{I,fg} = - \sum_{x \in \mathcal{X}} \left[ y(x) \log\big(p_{h}(x)\big) + \big(1 - y(x)\big) \log\big(1 - p_{h}(x)\big)\right]\,
\end{equation}
\begin{equation}
\label{eq:lovasz}
        L_{I, mIoU} = \sum_{x \in \mathcal{X}} \frac{1}{2} \overline{\Delta J_0}\big(m_0(p_h(x)\big) +  \overline{\Delta J_1}\big(m_1(p_h(x)\big)\,
\end{equation}
The mean IoU loss $L_{I,mIoU}$ considers the match between predicted and ground-truth human segmentation masks using the Lovász-Softmax formulation~\cite{berman2018lovasz}, which  penalizes incomplete binary segmentation masks, as defined in Equation~\ref{eq:lovasz}, where $\overline{\Delta J_i}$ is the Jaccard index for class $i$ and $m_i\big( \,p_h(x) \big)$ are the pixel errors for class $i$, both as specified in~\cite{berman2018lovasz}.
The loss function for IG-MHP combines these terms as follows:
\begin{equation}
\label{eq:ig-mhp}
        L_{IG-MHP} = \lambda L_{I,fg} + (1 - \lambda)L_{I,mIoU}\,.
\end{equation}
This encourages the network to produce complete and accurate human instance segmentation masks, focusing on precise boundaries and resolving ambiguities between foreground and background caused by overlapping instances.

\subsection{Multi-View Instance-Guided Multi-Human Parsing}
\label{sect:MVIG-MHP}
The second stage builds on IG-MHP by introducing a multi-view consistency loss, which utilizes 3D information to resolve ambiguities caused by overlapping human instances by considering the other viewpoints available.
The multi-view loss is formulated to enforce both instance identity consistency, ensuring that corresponding human instances across views are consistently segmented, and body part consistency, encouraging the model to maintain accurate body part segmentation across views.

Consider a 3D scene composed of a set $\mathcal{P}$ of sparse 3D points, and $N$ views of the same 3D scene. Let $p_j^i$ be the projection of point $P_j$ on the i-th view. The instance identity consistency loss $L_{identity}$ ensures that, for a sparse set of points $\{P_j\}$ all projections $p_j^i$ are associated with the same human instance as determined by the 3D ground truth, as formulated here:
\begin{equation}
    L_{identity} = \sum_{i=1}^{N} \sum_{p_j \in \mathcal{P}} \left[ y(p_j^i) \log\big(p_{h}(p^i_j)\big) + \big(1 - y(p_j^i)\big) \log\big(1 - p_{h}(p_j^i)\big)\right]
\end{equation}
Similarly, predictions of body parts from each view should also be consistent across views. The body part consistency loss $L_{part}$ aggregates part predictions from multiple views to refine and provide sparse supervision to body part segmentation. For each 3D point in $\mathcal{P}$, the optimal body part label $c^*$ is estimated by aggregating all contributions from the $N$ views, considering the label score $p(p_j^i|c)$ given by the network as confidence:
\begin{equation}
    c^* = arg\,\max\limits_{c \in C}\ \sum_{i=1}^{N} p(\,p_j^i\,|c\,)\,,
\end{equation}
\begin{equation}
\label{eq:ce-part}
     L_{part} = \sum_{i=1}^{N} \sum_{p_j \in \mathcal{P}_{\beta}} \left[ c^* \log\big(p_{i}(p^i_j)\big) + (1 - c^*) \log\big(1 - p_{i}(p_j^i)\big)\right]
\end{equation}
where $C$ represents the set of body part labels.
The aggregated label $c^*$ is then propagated back to each view, considering the cross-entropy loss between $c^*$ and the predicted label in each view, as in Equation~\ref{eq:ce-part}.
As the projection of a 3D point belonging to an occluded region could contribute to the loss with the wrong label, a projected 3D point is included in the loss only if it is sufficiently close to the visible surface (within a threshold $\beta$). The set of such filtered points is denoted as $\mathcal{P}_{\beta}$ in Equation~\ref{eq:ce-part}.
Such threshold on points distance has been set to 30\,cm, taking into account the average ``thickness'' of the human body.
The multi-view consistency loss $L_{MV}$ is then:
\begin{equation}
        L_{MV} = L_{identity} + L_{part}\,,
\end{equation}
which combined with Equation~\ref{eq:ig-mhp} gives the loss of the MVIG-MHP framework:
\begin{equation}
        L_{MVIG-MHP} = L_{MV} + L_{IG-MHP}\,.
\end{equation}

\section{Panoptic-HuIS dataset}
\label{sect:Dataset}
As multi-view human instance segmentation datasets featuring heavily overlapped and occluded people are currently lacking in the literature, we introduce a novel annotation technique to generate accurate human instance segmentation masks from multi-view RGB+D images and 3D human skeletons. The method is applied to the CMU Panoptic Studio dataset~\cite{Joo_2017_TPAMI}, an extensive multi-view dataset capturing motion and social interactions in various activities. The capture system is composed of 480 VGA cameras, 31 HD cameras, and 10 Kinect sensors arranged around a dome.
In this study, we only focus on sequences framing multiple overlapped people (\emph{160224-haggling1}, \emph{160226-haggling1}, \emph{170407-haggling-a1}, \emph{170407-haggling-a2}, \emph{170407-haggling-a3}, and \emph{160422-ultimatum1}) from Kinects, giving RGB+D information. The new dataset we present, Panoptic-HuIS, includes 1360 frames annotated with human instances. Frames have been carefully selected considering multiple overlapped people. Even if the dataset is not large-scale, it is mainly intended to provide instance weak-supervision during model fine-tuning, so large volumes of data are not needed.

The proposed annotation technique takes advantage of point clouds and 3D body skeletons to produce human instance segmentation masks from multi-view video sequences. An overview of the pipeline is shown in the top section in Figure~\ref{fig:framework}. Given video sequences from Kinect sensors, scene point clouds are generated by merging the depth maps captured for each considered frame, followed by statistical outlier removal to refine noisy data. 
Each point is then annotated by associating it with the nearest 3D skeleton joint, enabling clear separation between multiple individuals in the scene.
To produce human segmentation masks from these annotations, 3D points are projected onto the image plane of each camera using the available camera calibration parameters. 
However, due to inherent sparsity and noise in the point clouds, these projections cannot directly form complete masks. Instead, seed points are extracted from sparse projections and used as prompts in a promptable segmentation model (e.g. Segment Anything~\cite{kirillov2023segany}), to obtain an accurate segmentation mask. To ensure seeds are evenly distributed over visible, non-occluded regions, the system employs K-means clustering to strategically distribute these seeds, adapting cluster counts based on point density per instance.
For challenging cases, such as point clouds that are particularly sparse in the legs region, 3D skeletal joints for knees and ankles are used as additional seed points.
When dealing with heavily occluded bodies, SAM often merges overlapping instances into a single mask. To mitigate this effect, instances are segmented sequentially, 
from farthest to nearest the camera, exploiting depth information to resolve ambiguities. Masks are finally refined by re-feeding them into SAM to improve accuracy.

\begin{figure}[tb]
    \centering
    \includegraphics[width=0.99\textwidth]{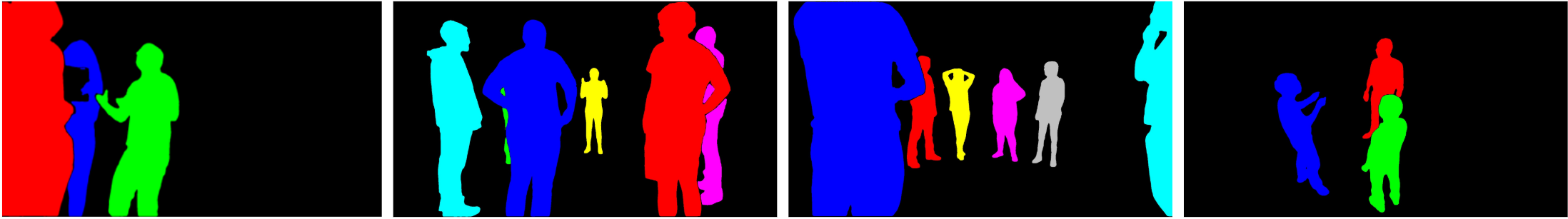}
    \caption{Sample images with human instance segmentation annotations from the Panoptic-HuIS dataset.}
    \label{fig:annotations}
\end{figure}

This approach ensures the generation of robust and coherent multi-view human instance segmentation masks even in complex, occluded scenarios.
Samples from the generated dataset are shown in Figure~\ref{fig:annotations}.

\section{Experiments}

\subsection{Evaluation of Multi-Human Parsing under Occlusion Scenarios}
To investigate the impact of overlapping human instances on the performance of state-of-the-art multi-human parsing models, we evaluated different architectures under varying levels of occlusion. Specifically, we considered two top-down approaches, RP R-CNN~\cite{yang2020renovating} and AIParsing~\cite{zhang2022aiparsing}, the bottom-up M2FP~\cite{cheng2022masked} and the single-stage approach Uniparser~\cite{chu2024uniparser}. The models are tested on the CIHP dataset~\cite{gong2018instance}, focusing on subsets of the validation split categorized by the degree of overlap between individuals.
The degree of overlap for each image is defined as the maximum Intersection-over-Union (IoU) computed between human ground-truth bounding boxes. The CIHP official validation split is then divided into four subsets: CIHP-O20, CIHP-O40, CIHP-O60, and CIHP-O80, containing images with an overlap of at least 20\%, 40\%, 60\%, and 80\%, respectively. Each model is evaluated on these subsets using two metrics: mean Intersection-over-Union (mIoU) on body parts, which serves as a global-level metric, and Average Precision based on Parts ($AP^p_{vol}$), that considers mean part IoU within a human instance. Results are summarized in Table~\ref{tab:overlap_analysis}. 
The evaluation confirms that parsing performance consistently deteriorates as the degree of overlap increases. Among the models, M2FP outperforms the two top-down approaches on the overall validation set and maintains its advantage under occlusion scenarios. However, it still experiences a significant performance drop, particularly in images with 80\% instance overlap, where its performance decreases by 16.1\% in mIoU. We can observe a similar behavior for the Uniparser network.
Comparing the two top-down approaches, RP R-CNN demonstrates a slower degradation in parsing quality than AIParsing. While AIParsing achieves better performance on the entire validation set, its accuracy deteriorates more rapidly with increasing overlap and performs the worst on CIHP-O80.
M2FP and Uniparser exhibit a less pronounced decline in instance-level segmentation performance, suggesting that, while occlusions remain challenging for bottom-up and single-stage approaches, they are less significant than for top-down methods.

\subsection{Evaluation of the Proposed Approach}

\renewcommand{\arraystretch}{1.3}
\begin{table}[tb]
\centering
\caption{Performance comparison of MHP models RP R-CNN, AIParsing, M2FP and UniParser on CIHP dataset and its subsets with varying degrees of human overlap.}
\label{tab:overlap_analysis}
\resizebox{1\textwidth}{!}{%
\begin{tabular}{lccccccccccccccccccc}
\toprule
         & \multicolumn{4}{c}{M2FP~\cite{yang2023deep}} & & \multicolumn{4}{c}{UniParser~\cite{chu2024uniparser}}  &  & \multicolumn{4}{c}{RP R-CNN~\cite{yang2020renovating}}  &  & \multicolumn{4}{c}{AIParsing~\cite{zhang2022aiparsing}} \\ \cline{2-5} \cline{7-10} \cline{12-15} \cline{17-20} 
         & \multicolumn{1}{r}{$mIoU_{p}$} & \multicolumn{1}{r}{$\Delta_{IoU}$} & \multicolumn{1}{r}{$AP^p_{vol}$} & \multicolumn{1}{l}{$\Delta_{AP}$} &  & \multicolumn{1}{r}{$mIoU_{p}$} & \multicolumn{1}{r}{$\Delta_{IoU}$} & \multicolumn{1}{r}{$AP^p_{vol}$} & \multicolumn{1}{l}{$\Delta_{AP}$} &      & \multicolumn{1}{r}{$mIoU_{p}$} & \multicolumn{1}{r}{$\Delta_{IoU}$} & \multicolumn{1}{r}{$AP^p_{vol}$} & \multicolumn{1}{l}{$\Delta_{AP}$} & & \multicolumn{1}{r}{$mIoU_{p}$} & \multicolumn{1}{r}{$\Delta_{IoU}$} & \multicolumn{1}{r}{$AP^p_{vol}$} & \multicolumn{1}{l}{$\Delta_{AP}$} \\ \hline
CIHP & 68.01 & - & 62.09 & - &  & 63.23 & - & 60.40 & - &  & 60.10 & - & 59.50 & - &  & 60.77 & - & 60.50 & - \\
CIHP-O20 & 66.50 & -2.2\% & 59.21 & -4.6\% &  & 61.77 & -2.3\% & 56.20 & -7.0\% &  & 58.54 & -2.6\% & 56.69 & -4.7\% &  & 58.19 & -4.3\% & 57.73 & -4.6\% \\
CIHP-O40 & 64.72 & -4.8\% & 58.60 & -5.6\% &  & 60.60 & -4.2\% & 54.80 & -9.3\% &  & 57.32 & -4.6\% & 54.60 & -8.2\% &  & 55.42 & -8.8\% & 56.00 & -7.4\% \\
CIHP-O60 & 63.83 & -6.1\% & 57.06 & -8.1\% &  & 59.87 & -5.3\% & 52.20 & -13.6\% &  & 55.35 & -7.9\% & 49.08 & -17.5\% &  & 50.82 & -16.4\% & 51.56 & -14.8\% \\
CIHP-O80 & 57.04 & -16.1\% & 57.53 & -7.3\% &  & 52.25 & -17.4\% & 50.80 & -15.9\% &  & 46.89 & -22.0\% & 48.70 & -18.2\% &  & 46.81 & -23.0\% & 50.36 & -16.8\% \\ \bottomrule
\end{tabular}
}
\end{table}

\begin{figure}[b]
    \centering
    \includegraphics[width=1\textwidth]{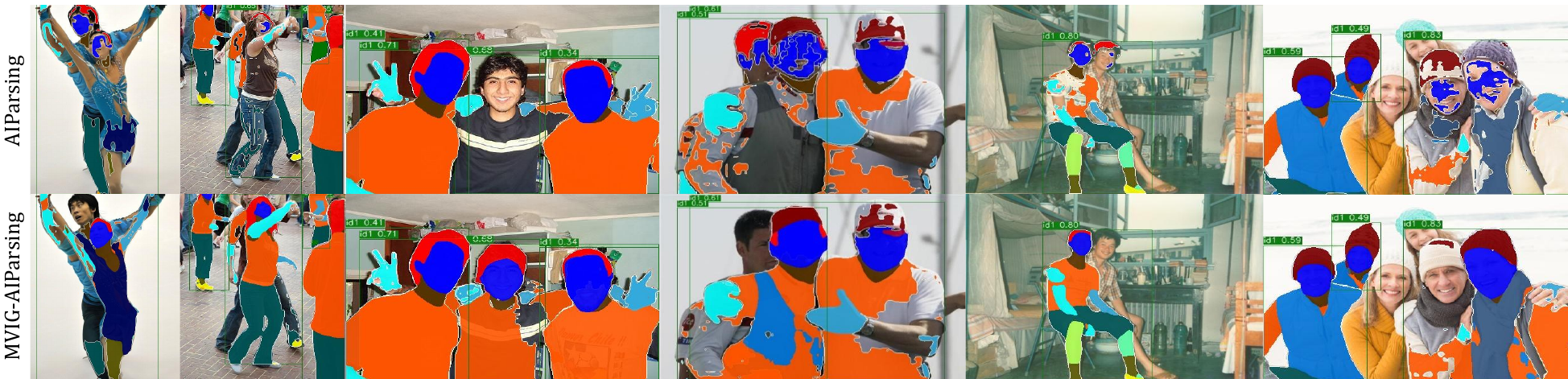}
    \caption{Qualitative comparison between AIParsing and MVIG-AIParsing.}
    \label{fig:results}
\end{figure}

\renewcommand{\arraystretch}{1.3}
\begin{table}[t]
\centering
\caption{Performance comparison of AIParsing, IG-AIParsing, and MVIG-AIParsing on the CIHP dataset and subsets. Bold and underlined values indicate respectively the best and second best performance for each metric within a given subset.}
\label{tab:results}
\resizebox{0.7\textwidth}{!}{%
\begin{tabular}{lllccclcccl}
\toprule
 &  &  & $mIoU_{p}$ & $mIoU_{p,m}$ & $mIoU_{p,ig}$ & $mIoU_{h,i}$ & $mIoU_{h}$ & $acc_{pixel}$ & $acc_{mean}$ & $AP^p_{vol}$ \\ \hline
\multirow{3}{*}{CIHP} &  & AIParsing & \textbf{60.77} & \textbf{64.74} & \textbf{65.69} & \textbf{70.40} & \textbf{90.05} & \textbf{90.29} & \textbf{71.37} & \textbf{60.50} \\
 &  & IG-AIParsing & 59.35 & 63.86 & 64.72 & {\ul 70.20} & 89.89 & 89.80 & 70.54 & 57.45 \\

\rowcolor{blue!8}  &  &  MVIG-AIParsing & {\ul 59.74} & {\ul 64.62} & {\ul 65.50} & {\ul 70.20} & {\ul 89.95} & {\ul 89.84} & {\ul 70.63} & {\ul 57.99} \\ \hline

\multicolumn{1}{c}{\multirow{3}{*}{CIHP-O20}} &  & AIParsing & \textbf{58.19} & {\ul 61.63} & {\ul 62.59} & 71.00 & 88.41 & \textbf{88.68} & 68.32 & \textbf{57.73} \\
\multicolumn{1}{c}{} &  & IG-AIParsing & 57.70 & 61.24 & 62.05 & {\ul 71.20} & {\ul 88.85} & 88.55 & {\ul 68.49} & 55.98 \\
\rowcolor{blue!8} \multicolumn{1}{c}{} &  & MVIG-AIParsing & {\ul 57.78} & \textbf{61.83} & \textbf{62.89} & \textbf{71.20} & \textbf{88.96} & {\ul 88.63} & \textbf{68.51} & {\ul 56.33} \\ \hline

\multicolumn{1}{c}{\multirow{3}{*}{CIHP-O40}} &  & AIParsing & 55.42 & 59.20 & 60.00 & 70.80 & 86.52 & 87.57 & 64.52 & \textbf{56.01} \\
\multicolumn{1}{c}{} &  & IG-AIParsing & {\ul 55.46} & {\ul 59.50} & {\ul 60.33} & {\ul 70.90} & {\ul 87.49} & \textbf{87.69} & \textbf{65.78} & 53.22 \\
\rowcolor{blue!8} \multicolumn{1}{c}{} &  &  MVIG-AIParsing & \textbf{55.50} & \textbf{59.79} & \textbf{60.40} & \textbf{71.00} & \textbf{87.57} & {\ul 87.66} & {\ul 65.63} & {\ul 53.74} \\ \hline
\multirow{3}{*}{CIHP-O60} &  & AIParsing & 50.82 & 54.81 & 55.96 & 68.10 & 80.75 & 84.84 & 59.47 & \textbf{51.56} \\
 &  & IG-AIParsing & {\ul 52.57} & {\ul 56.11} & {\ul 57.32} & {\ul 68.50} & {\ul 83.13} & {\ul 85.51} & \textbf{62.49} & 48.75 \\
\rowcolor{blue!8} &  & MVIG-AIParsing & \textbf{52.96} & \textbf{56.25} & \textbf{57.50} & \textbf{69.00} & \textbf{83.14} & \textbf{85.60} & {\ul 62.47} & {\ul 49.65} \\ \hline
\multirow{3}{*}{CIHP-O80} &  & AIParsing & 46.81 & 48.92 & 50.64 & 66.00 & 79.21 & 83.08 & 58.25 & \textbf{50.36} \\
 &  & IG-AIParsing & {\ul 47.61} & {\ul 50.08} & {\ul 51.75} & {\ul 67.00} & {\ul 81.97} & {\ul 83.95} & {\ul 60.03} & 47.54 \\
 \rowcolor{blue!8} &  &  MVIG-AIParsing & \textbf{48.49} & \textbf{51.15} & \textbf{52.70} & \textbf{67.90} & \textbf{82.68} & \textbf{84.19} & \textbf{60.59} & {\ul 48.48} \\ \bottomrule
\end{tabular}
}
\end{table}

\renewcommand{\arraystretch}{1.3}
\begin{table}[t]
\centering
\begin{minipage}{0.47\textwidth}
\centering
\caption{Performance comparison of MVIG-AIParsing on the CIHP dataset and its subsets with varying numbers of camera views during fine-tuning.}
\label{tab:camera_analysis}
  \resizebox{1\textwidth}{!}{%
  \begin{tabular}{lccccccccccc}
    \toprule
     & \multicolumn{2}{c}{CIHP-O20} &  & \multicolumn{2}{c}{CIHP-O40} &  & \multicolumn{2}{c}{CIHP-O60} &  & \multicolumn{2}{c}{CIHP-O80} \\ \cline{2-3} \cline{5-6} \cline{8-9} \cline{11-12} 
     & \multicolumn{1}{r}{$mIoU_p$} & \multicolumn{1}{r}{$mIoU_h$} &  & \multicolumn{1}{r}{$mIoU_p$} & \multicolumn{1}{r}{$mIoU_h$} &  & \multicolumn{1}{r}{$mIoU_p$} & \multicolumn{1}{r}{$mIoU_h$} &  & \multicolumn{1}{r}{$mIoU_p$} & \multicolumn{1}{r}{$mIoU_h$} \\ \hline
    2 views & 57.05 & 88.57 &  & 55.04 & 87.26 &  & 52.27 & 82.71 &  & 47.10 & 81.69 \\
    4 views & \textbf{57.78} & \textbf{88.96} &  & \textbf{55.50} & \textbf{87.57} &  & \textbf{52.96} & \textbf{83.14} &  & \textbf{48.49} & \textbf{82.68} \\
    8 views & 57.47 & 88.72 &  & 55.23 & 87.33 &  & 51.83 & 82.51 &  & 47.30 & 80.68 \\ \bottomrule
    \end{tabular}%
    }
\end{minipage}%
\hspace{0.2cm}
\begin{minipage}{0.47\textwidth}
\centering
\caption{Performance comparison of MVIG-AIParsing on CIHP subsets varying the distance threshold used for 3D point filtering.}
\label{tab:ablation}
  \resizebox{1\textwidth}{!}{%
    \begin{tabular}{lccccccccccc}
        \toprule
        & \multicolumn{2}{c}{CIHP-O20} &  & \multicolumn{2}{c}{CIHP-O40} &  & \multicolumn{2}{c}{CIHP-O60} &  & \multicolumn{2}{c}{CIHP-O80} \\ \cline{2-3} \cline{5-6} \cline{8-9} \cline{11-12} 
        & \multicolumn{1}{r}{$mIoU_{p}$} & \multicolumn{1}{r}{$mIoU_{h}$} &  & \multicolumn{1}{r}{$mIoU_{p}$} & \multicolumn{1}{r}{$mIoU_{h}$} &  & \multicolumn{1}{r}{$mIoU_{p}$} & \multicolumn{1}{r}{$mIoU_{h}$} &  & \multicolumn{1}{r}{$mIoU_{p}$} & \multicolumn{1}{r}{$mIoU_{h}$} \\ \hline
        20\,cm & 57.38 & 88.85 &  & 55.25 & \textbf{87.69} &  & 52.43 & 83.19 &  & 47.49 & 82.38 \\
        30\,cm & \textbf{57.78} & \textbf{88.96} &  & \textbf{55.50} & {\ul 87.57} &  & \textbf{52.96} & {\ul 83.24} &  & \textbf{48.49} & \textbf{82.68} \\
        40\,cm & 57.52 & 87.67 &  & 55.40 & 87.56 &  & 52.63 & \textbf{83.33} &  & 47.67 & 82.37 \\ \bottomrule
        \end{tabular}%
}
\end{minipage}
\end{table}

As our method aims at improving multi-human parsing in scenarios with occlusions, we focus on the top-down AIParsing architecture, which shows the highest drop in performance when dealing with severe occlusions (e.g., CIHP-O80). We integrate our approach into this network, referring to this enhanced version as MVIG-AIParsing. We make use of the new dataset Panoptic-HuIS to perform a fine-tuning of the model that has been pre-trained on the MHP dataset CIHP. Fine-tuning only affects the parsing head, while the detection head is kept frozen. We employ Stochastic Gradient Descent as in the original network, with learning rate set to $3e^{-4}$ and batch size equal to 8, for a maximum of 20 epochs. For IG-MHP we use $\lambda=0.5$, which provided the best results during experiments. 
For MVIG-MHP, we select several combinations of 4 adjacent cameras out of the 10 available to capture diverse spatial viewpoints for each scene. Multi-view loss considers 50 3D points randomly sampled from the original point cloud to ensure computational efficiency.

In addition to part-aware mean Intersection-over-Union ($mIoU_p$), and the Average Precision on parts ($AP^p_{vol}$), to evaluate our approach we also consider part-agnostic mean Intersection-over-Union ($mIoU_h$), which focuses on mIoU for human segmentation, and instance-level human mIoU ($mIoU_{h,i}$), which evaluates human mIoU for each instance. Pixel accuracy and mean pixel accuracy are also used to investigate weak supervision effects. 
As fine-tuning is performed on Panoptic-HuIS, while evaluation is performed on CIHP, we also inspect the effect of the domain gap between the two. In particular, we note that CIHP includes a large variety of different clothes and accessories, (e.g., `dress', `scarf', `hat'), while Panoptic-HuIS  images include main body parts and few clothes categories, lacking such diverse items. We add two additional metrics: $mIoU_{p,ig}$ that evaluates body-part mIoU ignoring clothes and accessories that never appear in Panoptic-HuIS images, and $mIoU_{p,m}$ that considers a label mapping between the categories present in CIHP and those found in Panoptic-HuIS (e.g. the category `hat' is mapped into the category `hair').

Table~\ref{tab:results} presents the performance on CIHP subsets with increasing overlap levels of three models: (i) AIParsing, used as baseline, (ii) IG-AIParsing, integrating single-view instance-guidance (IG-MHP), and (iii) MVIG-AIParsing, integrating MVIG-MHP. 
The results indicate that as overlap increases from 40\% to 80\%, MVIG-AIParsing consistently outperforms both AIParsing and IG-AIParsing across most metrics, showing superior body-part and human instance segmentation accuracy.
In heavily occluded scenarios, the MVIG training framework can enhance performance even considering the entire set of CIHP body-part categories, despite the domain gap between Panoptic-HuIS and CIHP.
In images with little to no overlap, the effect of the domain gap becomes more visible, but considering the common categories between the two datasets, $mIoU_{p,ig}$ and $mIoU_{p,m}$ metrics show that the parsing quality actually remains comparable with respect to the baseline. This demonstrates that any performance drop is limited to body-part labels that are unseen, thus not weakly supervised, during fine-tuning. 
Under occlusions, MVIG-MHP improves the completeness of human body masks ($mIoU_h$) to such an extent that it not only compensates for the domain gap, but yields a noticeable boost in accuracy, even when considering the full set of fine-grained CIHP labels ($mIoU_p$).
This is clearly shown in Figure~\ref{fig:results}, where MVIG-AIParsing produces more accurate and complete human parsing masks. It recovers larger portions of occluded human bodies, while also correctly associating body parts to the correct individuals, thanks to multi-view consistency enforced during fine-tuning. 
MVIG-AIParsing demonstrates to be superior in handling complex occlusions, while remaining comparable in scenarios without occlusions, making it the most robust solution overall.

\subsection{Ablation Study}
We investigate the impact of various parameters of the proposed approach, specifically the number of views used for multi-view loss computation and the threshold value used to filter 3D points considered for identity and part consistency.
We conduct experiments with groups of 2, 4, and 8 neighboring views to examine the impact of varying the number of cameras on the same scene. Results are shown in Table~\ref{tab:camera_analysis}. We can see that the best results are obtained using 4 views. This is due to the fact that, while 2 views are not enough to enforce strong constraints on body-part and instance predictions, 8 views span a very large portion of the scene, introducing inconsistent part prediction for the same projected 3D point. 
For 3D points filtering, in Table~\ref{tab:ablation}, we show results using different values of the distance threshold $\beta$, namely 20, 30, and 40\,cm. Choosing 30\,cm as threshold appears to be the best solution, providing a good balance between the number of retained points for multi-view loss computation and point label consistency.

\section{Conclusion}
In this work, we presented a training framework for multi-human parsing that leverages multi-view weak instance supervision to improve segmentation accuracy in occluded scenarios. In addition to injecting information about human shape, we enforce identity consistency across views by integrating a novel multi-view loss. To support this method, we introduce a semi-automatic annotation strategy to generate human instance segmentation masks from multi-view datasets with RGB+D data and 3D human skeletons, such as the CMU Panoptic Studio dataset.
Our results demonstrate that, in complex occlusion scenarios, state-of-the-art models struggle with segmentation accuracy. Our approach significantly mitigates this issue, consistently outperforming the baseline method AIParsing on human overlaps, while remaining comparable without occlusions.
We identify a few limitations in our solution, primarily due to limited diversity in clothing and accessories within the fine-tuning data, as well as the absence of fine-grained labels. To address this, we plan to enrich our dataset with more diverse samples and support the generation of fine-grained annotations, further broadening the applicability of our approach. We also aim to validate the generalization of our framework by applying it to other MHP architectures.

%
%
\bibliographystyle{splncs04}
\bibliography{main}
\end{document}